\documentclass{article}

\usepackage[english]{babel}

\usepackage[letterpaper,top=2cm,bottom=2cm,left=3cm,right=3cm,marginparwidth=1.75cm]{geometry}

\usepackage{amsmath}
\usepackage{amssymb}
\usepackage{graphicx}
\usepackage{multirow}
\usepackage{subcaption}
\usepackage{booktabs}
\usepackage[colorlinks=true, allcolors=blue]{hyperref}

\title{Pitfalls of Administrative Censoring in Survival Models with Time-Indexed Inputs}
\author{
Yanqi Xu\textsuperscript{1},
Hui Dai\textsuperscript{2},
Carlos Fernandez-Granda\textsuperscript{1},
Krzysztof J. Geras\textsuperscript{3,4},
Yiqiu Shen\textsuperscript{3}\\[0.5em]
\small \textsuperscript{1}Center for Data Science,
New York University, \small \textsuperscript{2}University of Chicagoy\\
\small \textsuperscript{3}NYU Grossman School of Medicine,
\small \textsuperscript{4}Ataraxis AI 
}

\begin{document}
\maketitle

\begin{abstract}
Survival models can model time-to-event outcomes using partially observed data. They are widely used in clinical prediction including cancer risk, disease progression, treatment response, and mortality. Recent models often rely on rich inputs collected at a specific clinical encounter, such as medical images, laboratory tests, electronic health record snapshots, or sensor measurements. In large retrospective datasets, these inputs are usually collected over many calendar years. As a result, they may contain clues about when they were acquired, through changes in devices, protocols, documentation, patient mix, or clinical practice. This creates a potential failure mode when outcomes are observed only up to a fixed study end date. More recent records necessarily have less possible follow-up than older records. A model that can infer record date from the input may therefore learn to predict how much follow-up was available, rather than the patient’s true risk of experiencing the event. We call this failure mode \emph{administrative-cutoff leakage}. In this paper, we characterize when this leakage can occur, distinguish it from classical informative censoring and genuine temporal changes in risk, and propose practical ways to detect it. In simulations, we show that administrative-cutoff leakage can inflate fixed-horizon AUC and can also affect Harrell’s C-index under realistic follow-up patterns. We then demonstrate the same behavior in a real mammography cohort. These results motivate a simple design principle for survival prediction: for an $n$-year prediction task, the dataset should provide at least $n$ years of potential follow-up after the latest input date. Otherwise, the models may be subject to bias induced by administrative-cutoff leakage.
\end{abstract}

\section{Introduction}
Survival analysis is the standard statistical framework for modeling time-to-event outcomes~\cite{kaplan1958nonparametric, cox1972regression}, where the central question is not only whether an event will occur, but when it will occur. This distinction is central in medicine because many decisions are tied to clinically meaningful time horizons: whether a patient is likely to develop cancer within the next several years, whether a disease is likely to progress before the next follow-up interval, or whether a treatment is expected to delay recurrence, progression, or death. A fundamental practical constraint in these settings is that not all subjects experience the event of interest during the observation period. Some individuals leave follow-up before the event occurs, while others remain event-free when the study ends. These partially observed outcomes are referred to as \emph{censored} observations. By accounting for censored observations, survival models can use partially observed follow-up rather than discarding individuals whose final event status is unknown~\cite{harrell1996multivariable}. This capability has made time-to-event modeling widely used across clinical prediction tasks, including cancer risk stratification ~\cite{yala2021toward, mikhael2023sybil}, modeling disease progression and recurrence using endpoints such as progression-free or disease-free survival~\cite{fda2018clinical}, and estimating treatment-associated outcomes for individualized decision-making~\cite{katzman2018deepsurv}.

Recent advances in deep learning have substantially expanded survival modeling by enabling time-to-event prediction from high-dimensional and heterogeneous clinical data, including medical imaging, radiology reports, and electronic health records (EHRs)~\cite{lecun2015deep, wulczyn2020deep, kim2021radiology, thorsenmeyer2022discrete}. However, because clinical events are often rare, training such models typically requires large datasets assembled over long calendar periods. We refer to these inputs as \emph{time-indexed data}: data acquired or indexed at a specific reference time, such as an image acquisition date, laboratory measurement date, or EHR snapshot date. When datasets span long periods, medical data distributions may evolve because of changes in patient populations, clinical practice, imaging devices, documentation patterns, or health-system workflows, leading to temporal drift~\cite{quionero2009dataset, ji2023temporal}. As a result, temporal information may be implicitly encoded in the data. The same representational flexibility that makes these models powerful also makes them vulnerable to shortcut learning: rather than learning genuine biological risk factors, they may exploit temporally correlated signals that appear predictive under standard evaluation but do not reflect true disease risk~\cite{geirhos2020shortcut}. This form of temporal shortcut learning is especially consequential in time-to-event prediction, where both labels and observation windows are tied to the timing of future events.

In this paper, we characterize a structural failure mode in survival prediction with time-indexed data, which we call \emph{administrative-cutoff leakage}. In long-running clinical cohorts, follow-up is often administratively censored at a fixed data-extraction, registry, or study-end date. As a result, patients enters the study later therefore have less possible follow-up time than patients entered earlier. Meanwhile, time-indexed data may contain temporal signatures of changing hardware, clinical protocols, assays, documentation practices, or preprocessing pipelines. Administrative-cutoff leakage arises when a model infers the input's reference time from these signatures and uses it as a shortcut for the observation process. The resulting predictions may therefore reflect the amount of follow-up available, rather than who is biologically more likely to experience the event.

This leakage differs from familiar shortcut-learning problems in medical imaging. Many shortcuts arise from \emph{dataset composition}: a feature such as age, scanner identity, or institution is spuriously correlated with the label in a particular sampled cohort, and the association may be weakened by resampling or balancing~\cite{zech2018confounding, jabbour2020exploiting}. Administrative-cutoff leakage instead arises from the construction of survival labels under a fixed cutoff date. Because reference time determines how much follow-up is observable for each subject, any input-derived signal of reference time can become a proxy for the observation window. This creates a structural pathway from temporal signatures in the input to the observed survival outcome, even when those signatures carry no biological information about future event risk. Consequently, balancing observed attributes is insufficient. Mitigation requires addressing the follow-up and evaluation structure of the survival dataset.

In this work, we study administrative-cutoff leakage as a form of calendar-time bias in long-term risk prediction. We make four contributions. First, we formulate the structural mechanism: leakage can arise when reference time is recoverable from time-indexed inputs and potential follow-up depends on reference time because outcomes are observed only up to a shared cutoff. Second, we propose practical diagnostics to test whether this structure is present in a dataset and whether a trained model has learned the resulting date-aligned shortcut. Third, we use controlled simulations to characterize how this mechanism affects common survival evaluation metrics, including fixed-horizon AUC and Harrell's C-index. Fourth, we validate the mechanism in a real world mammography cohort, showing that extending follow-up can remove observation-induced date alignment.

\section{Related Work}
\subsection{Shortcut Learning in Medical Data}
A growing body of work documents how machine learning models in medicine can exploit signals that are correlated with the target label in the training data but are not part of the underlying biological signal. In medical imaging, such shortcuts have been linked to institution- and acquisition-specific artifacts, demographic imbalance, and patient or healthcare-process variables embedded in the image data~\cite{zech2018confounding, badgeley2019deep, jabbour2020exploiting, degrave2021ai, larrazabal2020gender, gichoya2022ai}. Similar concerns arise beyond imaging: models trained on structured EHRs, clinical notes, ECGs, and other time-indexed clinical data may exploit treatment patterns, documentation practices, resource-use proxies, acquisition conditions, or other artifacts of healthcare delivery rather than the intended disease signal~\cite{caruana2015intelligible, obermeyer2019dissecting, ongly2024shortcut}. More broadly, shortcut learning has been described as a pervasive failure mode of flexible models, in which models learn decision rules that perform well under standard evaluation but fail when the shortcut no longer transfers~\cite{geirhos2020shortcut}.

Together, these studies show that high-capacity models can exploit signals that are easier to learn than the intended clinical target, including institutional artifacts, demographic attributes, acquisition-device signatures, documentation cues, and healthcare-process variables. These shortcuts are often framed as dataset-composition problems: a nuisance feature becomes spuriously associated with a diagnostic label or clinical outcome in the sampled cohort, leading to inflated internal performance and poor robustness under distribution shift. Our setting differs because the target is a time-to-event outcome. The input's reference time may be recoverable from the data, and that same reference time determines how much follow-up is observable before the administrative cutoff. Thus, a model that detects when an input was acquired can also infer how much follow-up was available. This shifts the problem from balancing a spurious attribute in the sampled cohort to examining how follow-up and censoring shape the survival outcome.

\subsection{Survival Analysis Basics and Informative Censoring}
For an individual $i$, let $X_i \in \mathcal{X}$ denote a time-indexed input (e.g.\ an image, laboratory panel, or EHR snapshot acquired at a reference date $D_i$), and let $T_i$ denote the future event time measured from baseline. Let $C_i$ denote the censoring time of the subject, that is the time of outcome being unobservable, such as time of dropping out of the study, measured from baseline. The observed survival outcome is
$$
Y_i = \min(T_i, C_i), \qquad
\delta_i = \mathbb{1}\{T_i \le C_i\},
$$
where $\delta_i=1$ indicates an observed event at $Y_i=T_i$, and $\delta_i=0$ indicates censoring at $Y_i=C_i$.

A standard assumption of survival modeling is independent censoring conditional on the variables used by the model (Figure \ref{fig:causal-graphs}a)~\cite{kalbfleisch2002statistical, tsiatis2006semiparametric},
$$C_i \perp T_i \mid X_i.$$
This condition allows the observed outcome to be written in terms of probabilities about the event time, even though some event times are censored. Conditional on $X_i=x$, there are two observed cases.
If $\delta_i=1$, then $T_i=y$ and censoring did not occur before $y$; if $\delta_i=0$, then $C_i=y$ and the event has not occurred by $y$. Thus
\begin{align*}
\Pr(Y_i=y,\delta_i=1\mid X_i=x)
&= \Pr(T_i=y, C_i\ge y\mid X_i=x),\\
\Pr(Y_i=y,\delta_i=0\mid X_i=x)
&= \Pr(C_i=y, T_i>y\mid X_i=x).
\end{align*}
By conditional independence, the terms involving $C_i$ factor away from the terms involving $T_i$. With conditional independence assumption, it is safe to only solve the event-time part of the likelihood
$$
\Pr(Y_i,\delta_i) \propto \left\{\Pr(T_i=y\mid X_i=x)\right\}^{\delta_i}
\left\{\Pr(T_i>y\mid X_i=x)\right\}^{1-\delta_i}.
$$
Most survival analysis models, including Cox proportional-hazards models and discrete-time survival models, differ in how they parameterize event-time probabilities, but rely on the conditional independence assumption to separate the event process from the censoring process~\cite{cox1972regression, allison1982discrete, gensheimer2019scalable, kvamme2019continuous}. Classical informative censoring refers to a violation of $C_i \perp T_i \mid X_i$, usually because a latent factor affects both censoring $C$ and the event time $T$ (Figure~\ref{fig:causal-graphs}b).

\section{Problem Formulation}
\label{sec:problem-formulation}
In this section, we formalize how reference time can become a source of leakage in survival prediction with time-indexed data. We first state the precondition that the input contains enough temporal information for a model to recover. We then describe two mechanisms by which a shared administrative cutoff can make this recovered reference time predictive of the observed survival outcome, even when it carries no information about biological event risk. Finally, we distinguish these leakage mechanisms from event drift, a contrasting setting in which reference time is genuinely associated with the event process.

\begin{figure}[h!]
\centering
\includegraphics[width=\linewidth]{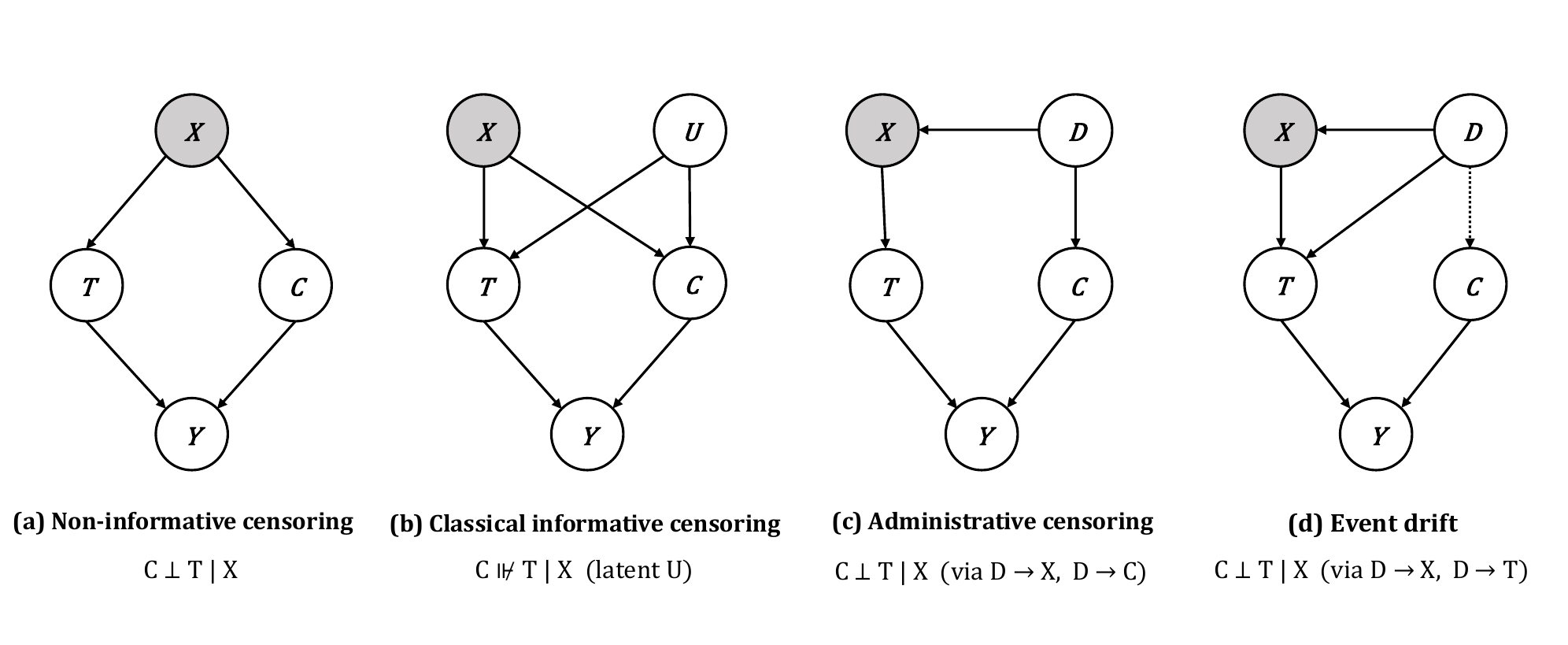}
\caption{\label{fig:causal-graphs} Causal structures for non-informative/informative censoring, administrative cenrsoring, and the event-drift positive control.}
\end{figure}

\paragraph{Temporal recoverability ($D\rightarrow X$).} Let $D_i$ denote the reference time associated with input $X_i$. Temporal recoverability holds when $D_i$ is predictable from $X_i$, i.e., when there exists a function $g$ such that $g(X_i) \approx D_i$. Equivalently, $X_i$ contains information about when it was acquired or indexed. We do not require perfect recovery of $D_i$; it is sufficient that the model can extract a signal correlated with reference time.

\begin{figure}[h!]
\centering
\includegraphics[width=\linewidth]{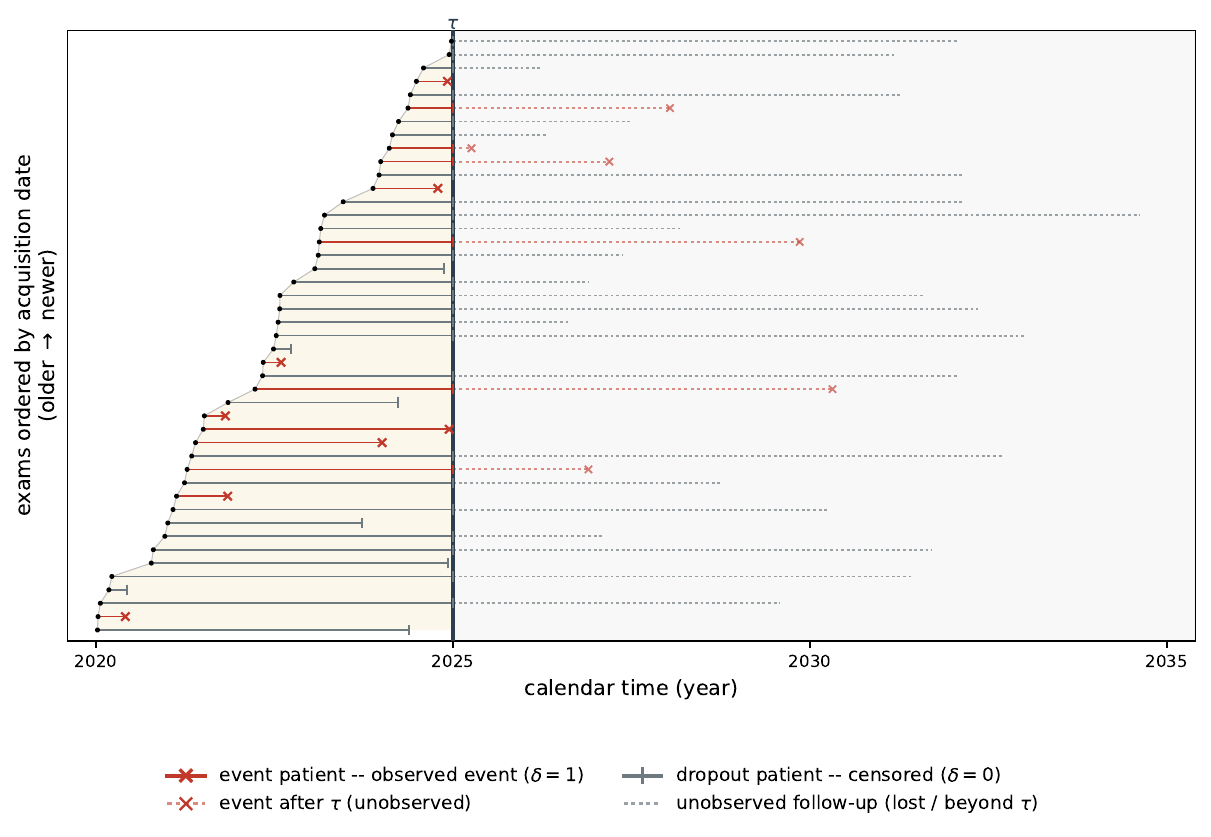}
\caption{\label{fig:condition2-timeline} Administrative cutoff geometry linking reference time \(D_i\) to potential follow-up \(A_i=\tau-D_i\).}
\end{figure}

\paragraph{Mechanism 1: Symmetric administrative censoring ($D\rightarrow C$).}
Many long-term cohorts are observed only up to a shared administrative cutoff time \(\tau\), such as a data-extraction date or registry end date. For an input indexed at reference time \(D_i\), the maximum observable follow-up duration is $$A_i = \tau - D_i.$$
Patients indexed later have smaller $A_i$, while patients indexed earlier have larger $A_i$ (Figure \ref{fig:condition2-timeline}). If $G_i$ denotes natural loss to follow-up, the observed censoring time is bounded by this administrative window:
$$
C_i = \min(G_i, A_i) = \min(G_i, \tau-D_i).
$$
For a prediction horizon $n$, only records with at least $n$ units of possible follow-up can be confirmed as event-free through that horizon, i.e., $A_i \ge n$, or equivalently $D_i \le \tau-n$. Records with $A_i<n$ can still have events observed before censoring, but if no event is observed, their event-free status through horizon $n$ remains unknown. Thus, the observability of horizon-$n$ labels depends on reference time $D$. If $D$ is recoverable from the input $X$, this creates a horizon-specific association between reference time $D$ and label observability/censoring $C$, even when the underlying event process $T$ does not depend on reference time (Figure~\ref{fig:causal-graphs}c).

\paragraph{Mechanism 2: Asymmetric administrative censoring ($D\rightarrow C$).}
Mechanism 2 retains the causal graph from Mechanism 1 (Figure~\ref{fig:causal-graphs}c), but considers a more realistic setting in which shortened follow-up affects observed events and observed non-events differently. Events such as cancer diagnosis or death can be recorded whenever they occur within the available observation window. In contrast, a long-horizon non-event can be established only through documented event-free follow-up, such as a follow-up visit, screening encounter, chart update, or registry linkage. Thus, for records without an observed event, the recorded censoring time may be the last documented event-free time $L_i \le \min(A_i,G_i)$, rather than the full administratively available window $\min(A_i,G_i)$.

This makes reference time more consequential for censoring among non-event records. Recently indexed records may still contribute observed early events, but they have had less calendar time to accumulate documented event-free follow-up, so $L_i$ tends to be shorter for newer $D_i$. Older records, by contrast, are more likely to have documented follow-up extending toward their natural loss-to-follow-up time $G_i$. This date-dependent difference in event and non-event ascertainment can distort not only horizon-specific AUC, but also comparable-pair and risk-set metrics such as Harrell's C-index.

\paragraph{Mechanism 3: Positive-control event drift ($D \rightarrow T$).}
Mechanism 3 is a positive control rather than an administrative-cutoff leakage mechanism. Here, reference time affects the latent event process itself (Figure \ref{fig:causal-graphs}d):
$$T_i \not\!\perp\!\!\!\perp D_i.$$
For example, disease incidence, screening sensitivity, treatment patterns, or diagnostic timing may change over calendar time. In this setting, date-associated prediction is not necessarily leakage: a model that recovers $D_i$ may be using information genuinely associated with event risk. We include this mechanism to distinguish observation-induced leakage from true temporal drift in the event process. We include S3 only as a positive control. A cohort-design remedy that removes observation-induced leakage should not eliminate signal that comes from true \(D\rightarrow T\) drift.

\section{Leakage Diagnosis}
\label{sec:quantifying_bias}

In this section, we describe diagnostics for detecting administrative-cutoff leakage at both the dataset and model levels. The dataset-level diagnostics examine whether the conditions for leakage are present: whether reference time is recoverable from the input, and whether recently indexed records lack enough follow-up to confirm long-horizon event-free status. At the model level, we ask whether a trained model appears to use this structure, by testing whether its risk scores follow the reference-time patterns induced by the administrative cutoff.

\subsection{Dataset-level diagnostics}

\paragraph{Temporal recoverability.}
We first test whether reference time is detectable from the input itself. Specifically, we train an auxiliary predictor $\phi$ to infer the reference time $D_i$ from the covariates $X_i$. Because $\phi$ is trained independently of the survival model and does not use survival outcomes, strong performance relative to a null or time-shuffled baseline indicates that reference-time information is present in the input representation and could be exploited by a flexible survival model.

\paragraph{Administrative cutoff rate.}
For each prediction horizon $n$, we quantify how many records are too close to the administrative cutoff to have their horizon-$n$ event-free status observed:
\begin{equation}
\rho_{\text{admin}}(n) =
\frac{|{i : D_i > \tau - n}|}{|{i : D_i \le \tau}|}.
\end{equation}
When $\rho_{\text{admin}}(n)>0$, some records were indexed within $n$ units of the administrative cutoff. These records may still have observed events before censoring, but they cannot be confirmed as event-free through horizon $n$, regardless of their underlying event risk.

\paragraph{Reference-time-only metric benchmark.}
To estimate how much apparent discrimination can arise from reference time alone, we use normalized $D_i$ as a risk score, with later reference times assigned higher risk. We compute
$$\mathrm{AUC}_n(D)=P(D_i>D_j\mid i\in\mathcal{P}_n,\; j\in\mathcal{N}_n),$$
and
$$C_{\text{Harrell}}(D)=\Pr(D_i>D_j \mid \delta_i=1,\; Y_i<Y_j),$$
where $\mathcal{P}_n=\{i:\delta_i=1,Y_i\le n\}$ denotes observed events by horizon $n$, and $\mathcal{N}_n=\{i:Y_i>n\}$ denotes records observed to be event-free beyond $n$. Above-chance values indicate that reference time alone can induce apparent predictive performance. Since $D_i$ contains no subject-specific biological risk information by construction, this benchmark quantifies how much a metric can be inflated by the observation structure induced by the administrative cutoff.


\subsection{Model-level leakage detection}
\label{sec:bias_metrics}

Dataset-level diagnostics establish whether administrative-cutoff leakage is possible. We next ask whether a trained model appears to have used this shortcut. The key idea is to compare records with the same observed event status and similar observed follow-up time. After this outcome matching, a model that is not using reference-time information should not systematically assign higher risk to the later-indexed record.
\paragraph{Outcome-Conditional Date Concordance (OCDC).}
Let $\hat{r}_n(i)$ denote the model's predicted $n$-year cumulative risk for record $i$, let $D_i$ denote its reference time, let $\delta_i$ denote the event indicator, and let $Y_i=\min(T_i,C_i)$ denote the observed time. For each horizon $n\in\{1,2,3\dots\}$ and tolerance parameter $w>0$, define the outcome-matched comparable set
$$
  \mathcal{C}_n =
  \bigl\{ (i,j) : \delta_i = \delta_j,\;
  |Y_i - Y_j| \le w,\;
  D_i > D_j \bigr\}.
$$
The per-horizon OCDC is
$$
  \text{OCDC}_n
  = \frac{1}{|\mathcal{C}_n|}
    \sum_{(i,j)\in\mathcal{C}_n}
    \Bigl[
      \mathbf{1}\!\left(\hat{r}_n(i) > \hat{r}_n(j)\right)
      + \tfrac{1}{2}\,
      \mathbf{1}\!\left(\hat{r}_n(i) = \hat{r}_n(j)\right)
    \Bigr].
$$
Thus, $\text{OCDC}_n$ estimates the probability that, among records with the same observed outcomes (same event status and similar observed follow-up time), the later-indexed record receives the higher predicted $n$-year risk. If the model is not using reference-time information, we expect $\text{OCDC}_n \approx 1/2$. In contrast, administrative-cutoff leakage can produce $\text{OCDC}n>1/2$, especially at longer horizons where the administrative cutoff rate $\rho{\mathrm{admin}}(n)$ is larger. However, elevated $\text{OCDC}_n$ is not by itself diagnostic of administrative-cutoff leakage: genuine temporal drift in the event process, as in Mechanism 3, can also induce date-aligned risk scores. We therefore interpret OCDC as a model-level warning signal, to be evaluated alongside dataset-level cutoff diagnostics and checks for event-rate drift.

\section{Mitigation Methods}
\label{sec:solutions}

Administrative-cutoff leakage can be reduced by breaking either of its two links: reference-time recoverability from the input, or the dependence of available follow-up on reference time. These correspond to representation-level and cohort-design remedies. We describe both, but evaluate only the cohort-design remedies in our experiments.

\paragraph{Reducing reference-time recoverability.}
One approach is to limit the model's ability to encode reference time, using techniques like preprocessing harmonization, domain-adversarial training, or representation-level debiasing~\cite{fortin2018harmonization,zhang2018mitigating}. These methods are attractive because they do not require additional follow-up collection. However, they offer no structural guarantee: residual date information may remain, and date-related features may be entangled with genuine biological signal. A weak date-recoverability probe can reduce concern, but cannot prove that all usable reference-time information has been removed.

\paragraph{Changing the follow-up structure.}
The more direct remedy is to make each evaluated horizon observable over the relevant reference-time window. For a horizon $n$, even the most recently indexed record must have at least $n$ years of possible follow-up:
$$
  \tau - \max_i D_i \ge n.
$$
Then every included record has at least $n$ years of potential follow-up, so $\mathcal{A}_n=\emptyset$ and $\rho_{\text{admin}}(n)=0$. f additional follow-up cannot be obtained, the same condition can be enforced by excluding records indexed too close to the cutoff, or by evaluating only shorter horizons satisfying $n \le \tau-\max_i D_i$.

\section{Controlled Simulation Study}
\subsection{Simulation Design}
\paragraph{Data-generating process.}
We simulate $N=10{,}000$ records over reference years $2018$--$2025$ with an administrative cutoff $\tau=2026$, repeating each experiment over $3$ seeds. Reference times $D_i$ are drawn uniformly across the study window and normalized to $t_i\in[0,1]$, where larger $t_i$ denotes a more recent record. Each record has $d=128$ features. To create controlled temporal recoverability, $16$ coordinates depend linearly on reference time $f_j(t_i)\propto t_i$, $16$ depend nonlinearly on reference time $f_j(t_i)\propto\sin(2\pi t_i)$, and the remaining $96$ are pure noise. Each date-dependent coordinate is generated as
$$
x_{ij}=\rho f_j(t_i)+\sqrt{1-\rho^2}\varepsilon_{ij},
\qquad
\varepsilon_{ij}\sim\mathcal{N}(0,1).
$$
Each date basis $f_j$ is standardized to zero mean and unit variance, so that $\mathrm{Var}(x_{ij})=1$ and $\mathrm{Corr}(x_{ij},f_j)=\rho$ exactly. Thus, $\rho$ controls how easily the record date can be recovered from the input: $\rho=0$ removes date information, while larger $\rho$ makes early and late records increasingly distinguishable.

The event times $T_i$ are generated from an exponential distribution with constant hazard $\lambda=0.07$, which yields an observed event rate of about $0.2$ under the cohort's censoring. Independent loss to follow-up is generated as $G_i\sim\mathrm{Exponential}(\text{mean}=10\text{ years})$. The observed censoring time is
$C_i=\min(\tau-D_i,,G_i),$
and an event is observed when $T_i\le C_i$. Therefore, later records have shorter observable follow-up only because they are closer to the administrative cutoff.

\paragraph{Simulation Scenarios.}

We compare the three scenarios described in Section~3. \textbf{S1} implements symmetric administrative censoring, with $C_i=\min(G_i,\tau-D_i)$. This creates date-dependent follow-up length because recent records are closer to the cutoff, but the censoring rule applies symmetrically to events and non-events. \textbf{S2} keeps the same event-time distribution as S1, but makes documented event-free follow-up shorter for recently indexed records and longer for earlier records. In both S1 and S2, the latent event time is independent of $X_i$ and $D_i$, so any above-chance discrimination must come from the observation process rather than true risk information. \textbf{S3}is a positive control in which event risk increases over reference time, so later records have genuinely higher event risk than earlier records:
$h_i=\lambda\exp\!\big(\beta\,(t_i-\tfrac{1}{2})\big),$
with $\beta=1.5$. In S3, date-associated performance reflects genuine event drift rather than administrative-cutoff leakage.

\paragraph{Models and evaluation.}
The cohort is split $70/30$ into train and test sets. We first train a date-recoverability probe $\phi$ to predict reference year from $X_i$. We then train three predictors using $X_i$ only: a multitask horizon-specific classifier, a discrete-time survival model~\cite{gensheimer2019scalable}, and a Cox proportional-hazards model~\cite{cox1972regression}. Performance is evaluated using horizon-specific $\mathrm{AUC}@n$~\cite{heagerty2005survival}, Harrell's C-index~\cite{harrell1996multivariable}, OCDC, and the administratively truncated fraction $\rho$. For mitigation, we compare no correction, inverse probability of censoring weighting (IPCW)~\cite{robins2000correcting}, extending the cutoff, and restricting the cohort to records with sufficient potential follow-up.

\subsection{Results}
\paragraph{Structural diagnostics.}
Table~\ref{tab:conditions} verifies the two structural ingredients needed for administrative-cutoff leakage. First, reference time becomes increasingly recoverable from the features as $\rho$ approaches $1$ (panel a). Second, potential follow-up decreases for later records, so the fraction of records that cannot be observed through a given prediction horizon increases with the horizon length (panel b). The reference-time-only benchmark (panel c) shows the resulting metric behavior before any outcome model is trained: S1 inflates horizon-specific AUC but not Harrell's C-index, S2 inflates both metrics through asymmetric event-free follow-up, and S3 produces date-only discrimination because calendar time truly affects event risk.

\begin{table}[h!]
\centering
\small
\setlength{\tabcolsep}{5pt}
\begin{subtable}[t]{0.39\linewidth}
\centering
\begin{tabular}{@{}ccc@{}}
\toprule
$\rho$ & Chance & Macro-AUC \\
\midrule
0.0 & 0.500 & 0.496 $\pm$ 0.003 \\
0.2 &       & 0.685 $\pm$ 0.005 \\
0.4 &       & 0.849 $\pm$ 0.002 \\
0.6 &       & 0.928 $\pm$ 0.002 \\
0.8 &       & 0.971 $\pm$ 0.002 \\
1.0 &       & 0.997 $\pm$ 0.001 \\
\bottomrule
\end{tabular}
\caption{\label{tab:cond2} Reference-time recoverability.}
\end{subtable}
\hfill
\begin{subtable}[t]{0.57\linewidth}
\centering
\begin{tabular}{@{}cccc@{}}
\toprule
$n$ & $\rho_{\text{admin}}(n)$ & AUC$_n(D)$ & $C(D)$ \\
\midrule
1\,yr & 0.000 & 0.503 $\pm$ 0.012 & \multirow{5}{*}{0.506 $\pm$ 0.003} \\
2\,yr & 0.146 & 0.569 $\pm$ 0.004 & \\
3\,yr & 0.290 & 0.621 $\pm$ 0.005 & \\
4\,yr & 0.434 & 0.679 $\pm$ 0.005 & \\
5\,yr & 0.574 & 0.745 $\pm$ 0.013 & \\
\bottomrule
\end{tabular}
\caption{\label{tab:cond1} Reference-time-only benchmarks.}
\end{subtable}
\vspace{0.75em}

\begin{subtable}[t]{\linewidth}
\centering
\begin{tabular}{@{}llccc@{}}
\toprule
Scenario & Model-free mechanism & $\rho_{\text{admin}}(5)$ & AUC$_5(D)$ & $C(D)$ \\
\midrule
S1 & symmetric administrative censoring & 0.574 & 0.745 $\pm$ 0.013 & 0.506 $\pm$ 0.003 \\
S2 & asymmetric event-free follow-up & 0.574 & 0.824 $\pm$ 0.005 & 0.610 $\pm$ 0.003 \\
S3 & positive-control event drift $D\to T$ & 0.574 & 0.831 $\pm$ 0.002 & 0.601 $\pm$ 0.003 \\
\bottomrule
\end{tabular}
\caption{\label{tab:scenario-diagnosis} Reference-time-only benchmarks across censoring scenarios.}
\end{subtable}
\caption{\label{tab:conditions} Dataset-level diagnostic checks before model training. Panels show that reference time is recoverable from the simulated inputs, that administrative truncation increases with horizon, and that reference time alone can produce metric inflation depending on the censoring scenario.}
\end{table}

\begin{figure}[h!]
\centering
\includegraphics[width=0.8\linewidth]{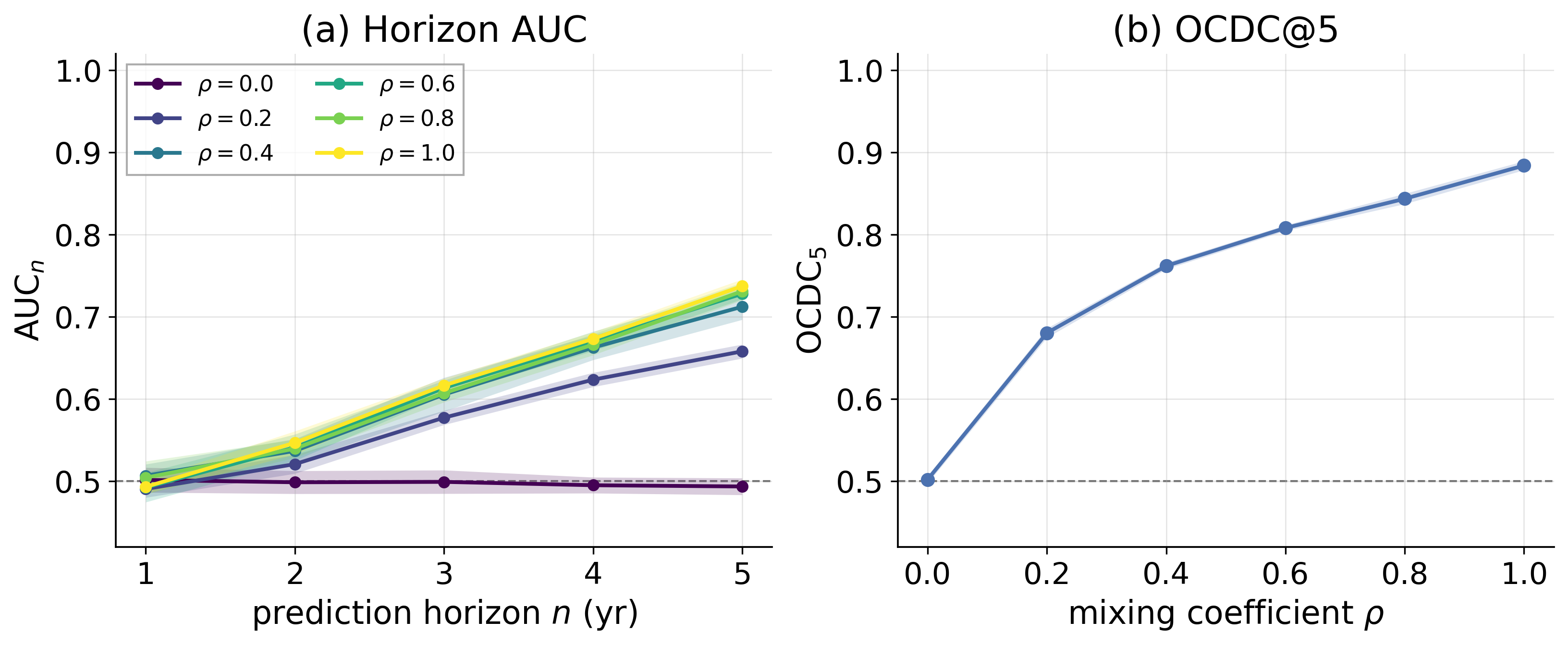}
\caption{\label{fig:sim-horizon} Model-level leakage under symmetric administrative censoring (S1). Horizon AUC and OCDC increase with feature--date recoverability and prediction horizon, showing that horizon-based models can convert input-derived reference time into inflated fixed-horizon performance even when the event process contains no biological signal.}
\end{figure}

\paragraph{Model evaluation.}
Under S1, Figure~\ref{fig:sim-horizon} shows that horizon-specific AUC increases as reference time becomes easier to recover from $X_i$ and as the prediction horizon becomes longer. When $\rho=0$, the model has no date information and performs near chance. When $\rho=1$, the model can identify more recent records and assign them higher risk. This inflates long-horizon AUC because recent non-event records often lack enough follow-up to enter the long-horizon control set. OCDC confirms that the model has learned this date shortcut: even among records with similar observed outcomes and follow-up time, later-indexed records receive higher predicted risk.

Table~\ref{tab:Scenarios} summarizes how this behavior differs across scenarios and models. In S1, $\mathrm{AUC}@5$ is inflated, but Harrell's C-index remains near chance because the comparable-pair structure is still symmetric. In S2, recent event-free records are documented with shorter follow-up, creating date-dependent ordering among comparable pairs; as a result, both $\mathrm{AUC}@5$ and C-index increase, including for Cox. In S3, all metrics increase for the expected reason: later records truly have higher event risk. Thus, recoverable reference time alone can inflate fixed-horizon AUC under administrative censoring, whereas C-index inflation requires either asymmetric documented follow-up or true temporal drift in event risk.

\begin{table}[h!]
\centering
\small
\setlength{\tabcolsep}{7pt}
\begin{tabular}{@{}lccc@{}}
\toprule
Scenario & AUC@5 & C-index & OCDC@5 \\
\midrule
\multicolumn{4}{@{}l}{\textbf{Multitask classification model}} \\
\midrule
S1        & 0.740 $\pm$ 0.022 & 0.505 $\pm$ 0.007 & 0.876 $\pm$ 0.004 \\
S2 (+asym.\ censoring)   & 0.821 $\pm$ 0.009 & 0.610 $\pm$ 0.004 & 0.923 $\pm$ 0.004 \\
S3 (+positive-control $D\!\to\!T$)  & 0.822 $\pm$ 0.004 & 0.596 $\pm$ 0.001 & 0.884 $\pm$ 0.006 \\
\addlinespace[0.35em]
\multicolumn{4}{@{}l}{\textbf{Discrete-time survival model}} \\
\midrule
S1        & 0.733 $\pm$ 0.020 & 0.506 $\pm$ 0.015 & 0.671 $\pm$ 0.028 \\
S2 (+asym.\ censoring)   & 0.806 $\pm$ 0.013 & 0.606 $\pm$ 0.004 & 0.856 $\pm$ 0.009 \\
S3 (+positive-control $D\!\to\!T$)  & 0.814 $\pm$ 0.011 & 0.587 $\pm$ 0.007 & 0.739 $\pm$ 0.023 \\
\addlinespace[0.35em]
\multicolumn{4}{@{}l}{\textbf{Cox proportional hazards}} \\
\midrule
S1        & 0.453 $\pm$ 0.049 & 0.505 $\pm$ 0.011 & 0.447 $\pm$ 0.062 \\
S2 (+asym.\ censoring)   & 0.820 $\pm$ 0.009 & 0.609 $\pm$ 0.002 & 0.931 $\pm$ 0.004 \\
S3 (+positive-control $D\!\to\!T$)  & 0.822 $\pm$ 0.005 & 0.597 $\pm$ 0.004 & 0.878 $\pm$ 0.007 \\
\bottomrule
\end{tabular}
\caption{\label{tab:Scenarios} Model performance at five years across censoring scenarios. Horizon-based models inflate AUC under symmetric administrative censoring (S1), whereas the continuous-time Cox model remains near chance in S1. Under asymmetric event-free follow-up (S2), date-dependent ordering affects AUC, C-index, and OCDC across model classes. S3 is a positive control in which reference time is genuinely predictive of event time.}
\end{table}

\subsection{Mitigation Results}

Table~\ref{tab:remedy} compares estimator-level reweighting with cohort-level remedies. IPCW has limited impact across three scenarios because it reweights only observed data; it cannot create missing five-year event-free controls for recent records. Therefore, $\rho_{\text{admin}}(5)$ remains $0.57$, and the inflated metrics change only modestly.

Cohort-level remedies target the source of leakage. Extending follow-up until $\tau-\max_i D_i\ge 5$, or restricting the cohort to records with $D_i\le\tau-5$, removes five-year administrative truncation and sets $\rho_{\text{admin}}(5)=0$. Under S1, both remedies return $\mathrm{AUC}@5$ to chance. Under S2, follow-up extension also removes the asymmetric-follow-up effect, reducing $\mathrm{AUC}@5$ from $0.806$ to $0.513$ and C-index from $0.606$ to $0.505$.

Restriction is less effective under S2 because it ensures sufficient potential follow-up but does not equalize documented event-free follow-up. Later records can still have shorter documented non-event follow-up, so $\mathrm{AUC}@5$ remains elevated, decreasing from $0.806$ to $0.630$. Thus, restriction attenuates asymmetric follow-up bias, whereas follow-up extension removes it in this simulation. In S3, residual discrimination after mitigation reflects true temporal drift in event risk, not observation-induced leakage.

\begin{table}[h!]
\centering
\small
\setlength{\tabcolsep}{6pt}
\begin{tabular}{@{}llccc@{}}
\toprule
Scenario & Remedy & $\rho_{\text{admin}}(5)$ & AUC@5 & C-index \\
\midrule
\multirow{4}{*}{S1} & None                         & 0.574 & 0.733 $\pm$ 0.020 & 0.506 $\pm$ 0.015 \\
                    & IPCW reweighting             & 0.574 & 0.704 $\pm$ 0.023 & 0.501 $\pm$ 0.019 \\
                    & Extend follow-up             & 0.000 & 0.497 $\pm$ 0.021 & 0.496 $\pm$ 0.017 \\
                    & Restrict cohort              & 0.000 & 0.500 $\pm$ 0.024 & 0.498 $\pm$ 0.025 \\
\midrule
\multirow{4}{*}{S2} & None                         & 0.574 & 0.806 $\pm$ 0.013 & 0.606 $\pm$ 0.004 \\
                    & IPCW reweighting             & 0.574 & 0.779 $\pm$ 0.022 & 0.621 $\pm$ 0.016 \\
                    & Extend follow-up             & 0.000 & 0.513 $\pm$ 0.017 & 0.505 $\pm$ 0.015 \\
                    & Restrict cohort              & 0.000 & 0.630 $\pm$ 0.008 & 0.539 $\pm$ 0.014 \\
\midrule
\multirow{4}{*}{S3} & None                         & 0.574 & 0.814 $\pm$ 0.011 & 0.587 $\pm$ 0.007 \\
                    & IPCW reweighting             & 0.574 & 0.787 $\pm$ 0.013 & 0.577 $\pm$ 0.012 \\
                    & Extend follow-up             & 0.000 & 0.581 $\pm$ 0.021 & 0.565 $\pm$ 0.015 \\
                    & Restrict cohort              & 0.000 & 0.532 $\pm$ 0.028 & 0.520 $\pm$ 0.026 \\
\bottomrule
\end{tabular}
\caption{\label{tab:remedy} Mitigation results. IPCW leaves the administratively truncated fraction unchanged and only modestly changes inflated metrics. Cohort-design remedies remove the structural S1 bias by making the five-year horizon observable for all retained records; under S2, follow-up extension is more effective than restriction because it also reduces date-dependent differences in documented event-free follow-up. S3 is retained as a positive control for true event drift.}
\end{table}

\section{Real World Dataset Experiments}
\subsection{Dataset}
We validate our findings on an internal full-field digital mammography (FFDM) screening cohort containing $25{,}511$ exams from $8{,}702$ women, acquired between January 2013 and August 2020. Cancer outcomes were identified from pathology records, and cancer-free follow-up was confirmed through subsequent screening encounters. 

The original outcome extraction ended in August 2020, so documented follow-up also stopped at that date. This naturally creates an S2 setting with asymmetric administrative censoring: later exams can contribute observed cancers, but have much shorter documented event-free follow-up than earlier exams. In the original cohort, median documented follow-up was $1.3$ years for later exams and $4.5$ years for earlier exams. We then extended outcome ascertainment through August 2025, giving every exam at least five years of potential wall-clock follow-up. The follow-up-extended cohort had median documented follow-up of $5.9$ years. We refer to the original cohort as V1 and the follow-up-extended cohort as V2. The two versions contain the same images and reference dates, but differ in the amount of documented follow-up.
Data were split at the woman level into $70\%$ training and $30\%$ test sets, preventing exams from the same woman from appearing in both partitions. The same woman-level split was used for V1 and V2.

\subsection{Experiment setup}
We trained two CNN models with identical architecture and optimization settings on datasets V1 and V2 respectively. Each model used a discrete-time survival head to predict breast cancer risk at horizons $n\in{2,3,4,5}$ years~\cite{kvamme2019continuous}. Both models are initialized from MammoCLIP weights~\cite{ghosh2024mammoclip} and trained with binary cross-entropy loss, batch size $8$, and the MammoCLIP hyperparameter setting on one NVIDIA A100 GPU. To assess date recoverability, we also trained a seperate model with the same backbone to predict the reference year from the input image.

At evaluation, positives at horizon $n$ are exams followed by malignant diagnosis within $n$ years, and controls are exams observed cancer-free beyond $n$ years. We compare V1 and V2 using administrative truncation $\rho_{\text{admin}}(n)$, date-only discrimination $\mathrm{AUC}_n(D)$ and $C(D)$, model discrimination $\mathrm{AUC}_n(\hat r)$ and $C(\hat r)$, and score-level date alignment $\mathrm{OCDC}_n$.

\subsection{Results}
The diagnostics show that V1 contains both ingredients for date-aligned leakage: reference time is recoverable from the image representation, and documented follow-up is strongly date-dependent (Table~\ref{tab:internal-followup-fix}). The reference-year probe achieved an AUC of $0.83$, indicating that acquisition time can be inferred from the images. As the prediction horizon increases, more exams lack sufficient documented follow-up: $\rho_{\text{admin}}(n)$ rises from $0.39$ at two years to $0.86$ at five years. Reference date alone is therefore strongly predictive of the evaluated label, with $\mathrm{AUC}_n(D)$ increasing from $0.75$ to $0.96$. The elevated date-only C-index, $C(D)=0.71$, further indicates that comparable pairs are systematically ordered by exam date under asymmetric event-free follow-up.

The model trained on V1 follows this date-dependent pattern despite not receiving acquisition date as an explicit input. Its horizon-specific AUC increases from $0.71$ to $0.86$, its C-index is $C(\hat r)=0.70$, and $\mathrm{OCDC}_n$ remains above chance ($0.56$--$0.58$). These results suggest that, in V1, the model score preserves date-aligned risk induced by the follow-up structure rather than purely reflecting image-based cancer risk. In V2, follow-up extension substantially reduces the date-dependent observability structure. After extension, $\rho_{\text{admin}}(n)$ remains near zero through five years ($\le0.07$), $\mathrm{AUC}_n(D)$ returns toward chance ($0.52$--$0.55$), and the date-only C-index is $C(D)=0.50$. The model trained on V2 also shows reduced date alignment: $\mathrm{AUC}_n(\hat r)$ becomes approximately constant across horizons ($\approx0.68$), $C(\hat r)=0.67$, and $\mathrm{OCDC}_n=0.49$. 

\begin{table}[h!]
\centering
\small
\setlength{\tabcolsep}{5pt}
\begin{tabular}{@{}ccccccc@{}}
\toprule
\multicolumn{7}{@{}l}{Date recoverability from input: 0.83} \\
\midrule
$n$ (yr) & $\rho_{\text{admin}}(n)$ & $\mathrm{AUC}_n(D)$ & $C(D)$ & $\mathrm{AUC}_n(\hat r)$ & $C(\hat r)$ & $\mathrm{OCDC}_n$ \\
\midrule
\multicolumn{7}{@{}l}{\textbf{Dataset V1}} \\
\midrule
2 & 0.39 & 0.75 & \multirow{4}{*}{0.71} & 0.71 & \multirow{4}{*}{0.70} & 0.56 \\
3 & 0.56 & 0.83 &  & 0.78 &  & 0.58 \\
4 & 0.73 & 0.91 &  & 0.84 &  & 0.58 \\
5 & 0.86 & 0.96 &  & 0.86 &  & 0.58 \\
\midrule
\multicolumn{7}{@{}l}{\textbf{Dataset V2}} \\
\midrule
2 & 0.00 & 0.52 & \multirow{4}{*}{0.50} & 0.68 & \multirow{4}{*}{0.67} & 0.49 \\
3 & 0.00 & 0.52 &  & 0.68 &  & 0.49 \\
4 & 0.00 & 0.53 &  & 0.68 &  & 0.49 \\
5 & 0.07 & 0.55 &  & 0.68 &  & 0.49 \\
\bottomrule
\end{tabular}
\caption{\label{tab:internal-followup-fix} Internal follow-up extension experiment. The same de-drifted test exams are evaluated under S2 asymmetric administrative censoring and after follow-up extension. Models are trained separately using the same architecture and training protocol. Date recoverability is measured once because the image set is unchanged across conditions.}
\end{table}

\section{Discussion}
This study identifies administrative-cutoff leakage as a structural failure mode in long-term risk prediction with time-indexed inputs. The failure requires two ingredients: reference time is recoverable from the input, and more recent records have less observable follow-up than earlier records. Together, these conditions allow models to use reference time as a shortcut for outcome observability rather than true event risk.

The simulation results show that administrative-cutoff leakage can bias both horizon-specific AUC and Harrell's C-index, with the affected metric depending on the censoring structure and learning algorithm. The internal mammography experiment supports this interpretation in a real imaging cohort. Without enough followup time, the collected data exhibit asymmetric administrative censoring (S2) which results in inflated per horizon AUC and C-index. Extending follow-up removed the date-only signal and reduced OCDC to chance. 

These findings highlight the importance of cohort design for long-term survival modeling. For an $n$-year prediction task, the most direct safeguard is to ensure at least $n$ years of potential follow-up after the latest input acquisition date, or to restrict evaluation to horizons observable for the included records.

\bibliographystyle{plain}
\bibliography{sample}

\end{document}